\documentclass[acmlarge,nonacm]{acmart}

\AtBeginDocument{%
  \providecommand\BibTeX{{%
    \normalfont B\kern-0.5em{\scshape i\kern-0.25em b}\kern-0.8em\TeX}}}

\usepackage{soul}
\usepackage{tabularx}
\usepackage{caption}
\usepackage{subcaption}
\usepackage{longtable}
\usepackage{array}
\usepackage{wrapfig,lipsum,booktabs}
\usepackage{textcomp}
\usepackage{multicol}
\usepackage{multirow}
\usepackage{comment}
\usepackage{amsmath}
\usepackage{enumitem,ragged2e}
\usepackage{xcolor,colortbl}
\usepackage{tabularx}
\newcommand{\AlgoName}{\textsc{MedInsight}\xspace}
\usepackage[most]{tcolorbox}
\usepackage[T1]{fontenc}
\usepackage{tgbonum}
\usepackage{caption}
\usepackage{booktabs,makecell}
\usepackage{indentfirst}

\usepackage[super]{nth}
\usepackage[ruled,vlined]{algorithm2e}
\SetKwInput{KwInput}{Input}
\SetKwInput{KwOutputput}{Output}
\SetKwInput{KwRequire}{Require}

\newtcolorbox{boxEnv}{
center,
left=0 mm,
top = 0.25 mm,
right = 0mm,
bottom =0.25 mm,
colframe=gray!90!black,
colback=black!5!white, 
boxrule=0.5pt,
}

\begin{document}

\title[\AlgoName]{\AlgoName: A Multi-Source Context Augmentation Framework for Generating Patient-Centric Medical Responses using Large Language Models} 

   \author{Subash Neupane, Shaswata Mitra, Sudip Mittal, Noorbakhsh Amiri Golilarz, Shahram Rahimi, Amin Amirlatifi} 
    \affiliation{ 
      \institution{Mississippi State University}
      \streetaddress{665 Perry St}
      \city{MSSTATE} 
      \state{MS} 
      \country{USA}
      \postcode{39762}
    }
    \email{sn922@msstate.edu,sm3843@msstate.edu, mittal@cse.msstate.edu, amiri@cse.msstate.edu, rahimi@cse.msstate.edu, amin@che.msstate.edu  }

\renewcommand{\shortauthors}{Neupane et al.}

\begin{abstract}

Large Language Models (LLMs) have shown impressive capabilities in generating human-like responses. However, their lack of domain-specific knowledge limits their applicability in healthcare settings, where contextual and comprehensive responses are vital. To address this challenge and enable the generation of patient-centric responses that are contextually relevant and comprehensive, we propose \AlgoName - a novel retrieval augmented framework that augments LLM inputs (prompts) with relevant background information from multiple sources. \AlgoName extracts pertinent details from the patient's medical record or consultation transcript. It then integrates information from authoritative medical textbooks and curated web resources based on the patient's health history and condition. By constructing an augmented context combining the patient's record with relevant medical knowledge, \AlgoName generates enriched, patient-specific responses tailored for healthcare applications such as diagnosis, treatment recommendations, or patient education. Experiments on the MTSamples dataset validate \AlgoName's effectiveness in generating contextually appropriate medical responses. Quantitative evaluation using the Ragas metric and TruLens for answer similarity and answer correctness demonstrates the model's efficacy. Furthermore, human evaluation studies involving Subject Matter Expert (SMEs) confirm \AlgoName's utility, with moderate inter-rater agreement on the relevance and correctness of the generated responses.
\end{abstract}





\maketitle

\section{Introduction}

In the healthcare domain, providing contextual and comprehensive medical information tailored to individual patients is crucial for enabling effective care. However, existing approaches often struggle to deliver personalized responses due to the distributed nature of medical data across multiple sources like patient records, medical literature, and online resources. While recent advances in Large Language Models (LLMs) have demonstrated their potential for understanding and communicating medical knowledge, their training objective of next-token prediction can lead to information loss, `memory distortion' \cite{peng2023check}, and the generation of plausible but incorrect content, known as hallucinations \citep{huang2023survey}. These shortcomings highlight the need for techniques to augment LLMs with contextually relevant information from diverse sources to ensure the delivery of reliable, patient-centric responses.


To address the challenge of adapting LLMs for specialized domains like healthcare, two main approaches have emerged: fine-tuning and augmenting the models with external knowledge. Fine-tuning involves further training a pre-trained LLM on domain-specific data to optimize its performance for targeted applications \cite{ovadia2023fine}. However, this method can be computationally expensive, limited by data availability, and susceptible to catastrophic forgetting, where the model forgets previously learned knowledge \cite{ovadia2023fine, kirkpatrick2017overcoming, goodfellow2013empirical, chen2020recall, luo2023empirical}. An alternative approach is In-Context Learning (ICL), which aims to enhance LLM effectiveness on new tasks by modifying the input prompts without changing the model weights \cite{chen2021meta, radford2019language, lampinen2022can}. A prominent implementation of ICL is Retrieval Augmented Generation (RAG) \cite{lewis2020retrieval, neelakantan2022text}, where information retrieval techniques are used to extract relevant knowledge from external sources and integrate it into the LLM's generated text. By augmenting the model's input with retrieved contextual information, RAG can adapt LLMs to domain-specific tasks without the drawbacks associated with fine-tuning.



While a patient's medical transcript captures their history and current condition, the background information required for comprehensive care, such as details about diseases, symptoms, diagnoses, and treatments, is often distributed across multiple sources like medical literature, clinical guidelines, and online knowledge bases. This fragmentation of relevant medical knowledge poses a significant challenge in providing personalized responses tailored to a patient's unique context. To address this challenge and effectively leverage the disparate sources of information, we propose MedInsight, a RAG framework capable of generating tailored medical responses by augmenting a patient's specific context from their transcript with pertinent background knowledge retrieved from various authoritative sources.

\begin{figure*}[h]
    \centering
    \includegraphics[scale=.65 ]{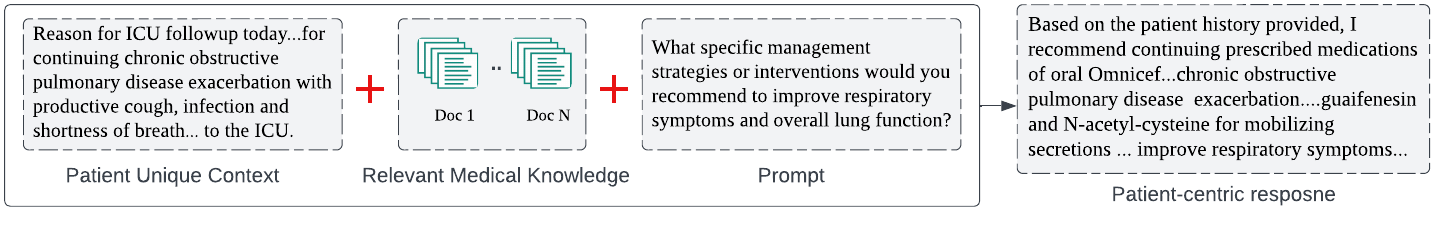}
    \caption{\AlgoName's context augmentation approach for generating patient-centric responses. Patient context from medical transcripts and prompt is augmented with relevant medical knowledge from authoritative sources into a comprehensive context input to the language model, enabling personalized patient-centric response generation.}
        \label{fig: overview_fig1_intro}
\end{figure*}

Generating patient-centric responses to medical queries requires augmenting the patient's context with relevant knowledge extracted from authoritative sources. Unlike tasks focused on specific facts, medical questions often necessitate a deeper understanding of relationships across multiple contexts spanning the patient's history, current symptoms, lab findings, and background medical knowledge. For instance, consider the question \textit{"For a patient experiencing a Chronic Obstructive Pulmonary Disease (COPD) exacerbation, what specific management strategies or interventions would you recommend to improve respiratory symptoms and overall lung function?"} Answering this effectively demands comprehending the overall patient context from their transcript, while also incorporating pertinent information about COPD exacerbations, treatment strategies, and respiratory management from medical knowledge sources. Figure \ref{fig: overview_fig1_intro} illustrates how MedInsight augments context from multiple sources to craft patient-centric responses. The prompt (query) is merged with relevant information extracted from the patient's medical transcript (i.e. patient unique context) and authoritative medical knowledge sources like textbooks. This formulates an augmented context combining the patient's details with related medical concepts, enabling MedInsight to generate a contextually relevant, personalized response tailored to the specific patient's needs.



By contextualizing patient information with medical knowledge, we empower patient and caregivers with the insights and tools necessary to enhance self care/patient care and optimize healthcare delivery. Our RAG approach generates patient-centric responses leveraging patient's unique context and extracting relevant medical knowledge based on patient's history for specific input. This enables both patient and caregivers to access relevant and personalized information, facilitating informed decision-making and improving patient outcomes.

The major contributions of this paper are:
\begin{itemize}
\item We demonstrate the possibility of tailoring medical recommendations for a specific patient need using patient context (medical transcript and health record). This equips caregivers with the knowledge and tools necessary to elevate patient care, enhance treatment results, and optimize the efficiency of healthcare delivery.

    \item We built a retrieval-augmented question-answering system that generates patient-centric responses leveraging patient's unique context and extracting medical knowledge based on patient's medical history for specific input.

\item We showcase and evaluated \AlgoName's proficiency in generating accurate and patient-centric responses through both qualitative and quantitative metrics. 
\end{itemize} 

The rest of the paper is organized as follows: Section \ref{background} discusses the background and related works, and section \ref{architecture} provides insight on \AlgoName's architecture and methodology. In Section \ref{experiment} we present our experiments, evaluation and discuss our results. Finally, Section \ref{conclusion} concludes the paper.

\section{Background and Related Works}
\label{background}

In this section, we commence by providing the foundational background of Large Language Models (LLM) and Retrieval Augmented Generation (RAG). Subsequently, we explore relevant works that have leveraged these methodologies within the healthcare domain.

\subsection{Large Language Models}
\label{large_language_models}

    Language models (LMs) can be referred to as computational models that have the ability to comprehend and generate human language. They learn to model the probability distribution of text, subsequently predict the likelihood of word sequences, or generate new text based on the input \citep{chang2023survey}. For instance, traditional language models like N-gram \citep{brown1992class} estimate the probability of a word by considering its context in preceding text. 
    
\begin{figure}
 \begin{subfigure}{0.36\textwidth}
     \includegraphics[width=\textwidth]{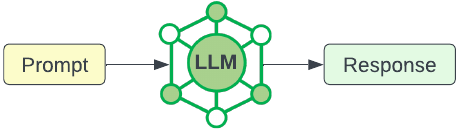}
     \caption{A simple LLM for response generation.}
    \label{fig:simple_llm}
 \end{subfigure}
 \hfill
 \begin{subfigure}{0.51\textwidth}
     \includegraphics[width=\textwidth]{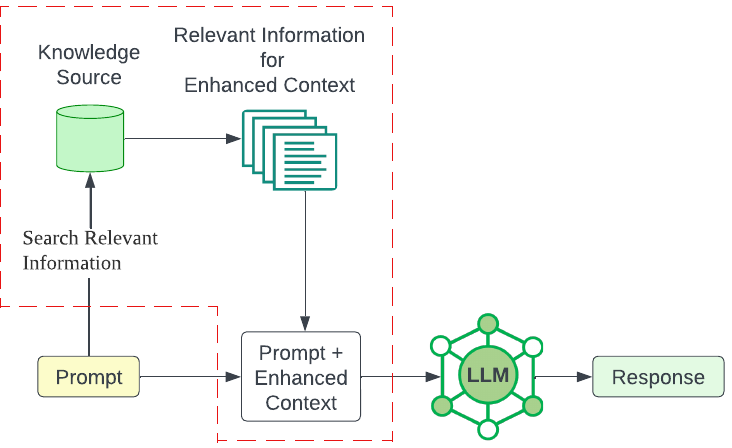}
     \caption{An example of simple RAG.}
    \label{fig:simple_rag}
 \end{subfigure}

  \vspace{3mm}
 \begin{subfigure}{0.51\textwidth}
     \includegraphics[width=\textwidth]{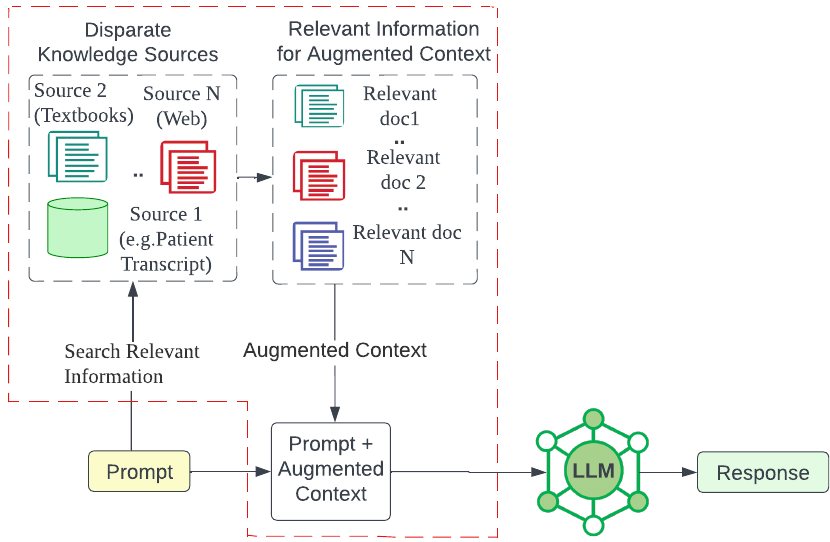}
     \caption{Our approach where we combine multi source context to generate personalized responses.}
    \label{fig:our_rag}
 \end{subfigure}
 \caption{The illustration emphasizes response generation using LLMs with and without external knowledge. FIg. \ref{fig:simple_llm} shows response generation without external knowledge,  Fig. \ref{fig:simple_rag} depicts a simple RAG that utilizes an single external source whereas Fig. \ref{fig:our_rag} depicts our approach where we combine multi source context to generate personalized response.}

\end{figure}

    Large Language Models (LLMs), on the other hand, are advanced language models trained on massive crawls of Internet text with massive parameter sizes and has exceptional learning capabilities. Recently, they have emerged as pivotal catalysts in several research domains including but not limited to Natural Language Processing (NLP) \citep{radford2019language, brown2020language}, cybersecurity \citep{mitra2024localintel}, and recommender systems \citep{zhang2023recommendation, hou2023large}. The state-of-the-art LLMs such as OpenAI GPT \citep{radford2019language}, Google’s PaLM \citep{ chowdhery2022palm}, Meta’s LLMA2 \cite{ touvron2023llama} primarily leverage the Transformer architecture \cite{ vaswani2017attention}.  Existing LLMs follow diverse transformer architectures and pre-training objectives such as employing solely decoders (as seen in GPT-2 and GPT-3), using only encoders (as exemplified by BERT~\cite{devlin2018bert} and RoBERTa \cite{liu2019roberta}), or adopting encoder-decoder structures (as seen in BART). Models can be trained through distinct methodologies: employing an \textit{autoregressive} approach, wherein the objective is to predict the subsequent word given the left-hand context; utilizing a \textit{masking} technique, akin to a fill-in-the-blanks problem, where the goal is to predict masked words with context on both sides; or adopting a strategy where the sequence is intentionally corrupted, followed by the task of predicting the original sequence. These models showcase exceptional proficiency in understanding and generating language, producing responses that closely emulate human expression and intention, as exemplified by the emergence of chatbot applications such as ChatGPT.  
   LLMs effectively apply their acquired knowledge and reasoning abilities to handle various downstream tasks including Named Entity Recognition (NER), text summarization, question-answering, and more. Figure \ref{fig:simple_llm} depicts a base LLM for these downstream tasks. This capability is attributed to inherent features of LLMs such as prompting or in-context learning \cite{brown2020language}, achieved through the provision of appropriate instructions or prompts \citep{zhu2023large}.

   \subsection{Retrieval Augmented Generation}
    Pre-trained LLMs, including models such as GPT and LLMA, have demonstrated their capabilities at acquiring thorough and detailed knowledge from training data. They leverage this acquired knowledge to generate text based on the provided input. Nevertheless, these models face significant limitations that impede wider deployment. For example, they may produce predictions that, while appearing plausible may not be non factual, commonly referred to as hallucinations \citep{huang2023survey}. This tendency is particularly pronounced when queries go beyond the model's training data or require current and updated information. To address this limitation, a promising approach is Retrieval Augmented Generation (RAG), introduced by \cite{lewis2020retrieval} in mid-2020. This method integrates external data retrieval into the generative process, thereby augmenting the model's capacity to offer precise and pertinent responses. An illustration of a simple RAG can be observed in Figure \ref{fig:simple_rag}.

   RAG comprises three essential elements: \textit{a knowledge database, a retriever}, and an \textit{LLM}. The knowledge database is capable of housing an extensive array of texts sourced from diverse outlets, tailored to the specific domain. For instance, in the medical domain, it might encompass information about medical condition (diseases), their symptoms, preventive measures, diagnosis and recommended medications, among other relevant content. The retriever employs a text encoder to compute an embedding vector for each text in the knowledge database. When presented with a user's query like \textit{``What is Kawasaki diseases and how does it impact my child’s health?''} the retriever utilizes the text encoder to output an embedding vector for the question 
   Depending on implementation retrieval component may be based Dense Passage Retrieval (DPR) \cite{karpukhin2020dense} and may follow a bi-encoder architecture. Next, a subset of texts, e.g., $k$, denoted as retrieved texts is extracted from the medical knowledge database with largest similarity, e.g., cosine similarity, to that of given question. Subsequently, these $k$ retrieved texts serve as the augmented context for the LLMs to generate an answer to the provided question.

\subsection{LLMs in Healthcare}



In the realm of Natural Language Processing (NLP), LLMs have sparked a revolution  with their outstanding performance \cite{bakker2022fine} in various tasks like summarization, question-answering, and Natural Language Generation (NLG) \cite{bubeck2023sparks}. Their versatile utility is prompting researchers to actively explore potential applications in the healthcare domain. This is evidenced by the success of ChatGPT in attaining a passing grade in United States Medical Licensing Examinations (USMLE) \cite{kung2023performance}. Additionally, a version of Med-PaLM2 that was fine-tuned using medical data has recently achieved state-of-the-art results, attaining the level of expertise demonstrated by human clinicians \cite{singhal2023towards}. 
Although LLMs have impressive capacities to generate human-like responses for different downstream tasks such as task-oriented question and answering, applying LLMs to medical domains is still challenging. This is because an LLM may lack comprehensive expertise in medical knowledge to which it has not been exposed. For instance, a model trained exclusively on texts written by William Shakespeare would struggle to perform well when queried about the symptoms of a disease \cite{ovadia2023fine}. 
In order to mitigate this issue, researchers are currently augmenting LLMs with external knowledge. Zakka et al., \cite{zakka2023almanac} introduced Almanac, a framework augmented with retrieval capabilities for medical guidelines and treatment recommendations. This framework was created to respond to 130 clinical questions formulated by a panel of five board-certified clinicians and resident physicians. The results demonstrated that Almanac surpassed GPT-4 in terms of factuality, safety, and correctness. This suggests that the incorporation of retrieval systems results in more precise and dependable responses to clinical inquiries.

Likewise, the authors in \cite{kang2023knowledge} presented a novel approach called KARD.  This method fine-tunes small LLMs to generate rationales obtained from language models with augmented knowledge retrieved from an external knowledge base or from a non-parametric memory. Lozano et al., \cite{lozano2023clinfo} on the other hand, introduced Clinfo.ai, an open-source workflow that incorporates end-to-end retrieval-augmented LLM chains. This workflow is specifically designed for querying, evaluating, and synthesizing medical literature into concise summaries to address questions on demand. To enhance the accuracy of Large Language Models (LLMs) such as GPT-3/4 on biomedical data, the authors in \cite{soong2023improving} employed retrieval methods. Subsequently, they qualitatively evaluated the performance of GPT-3.5 and GPT-4 against a custom RetA LLM using a set of 19 questions. The results indicated that LLMs when used alone exhibited more instances of hallucination compared to the customized approach. To incorporate general knowledge from LLMs into specific domains, Wang et al.,\cite{wang2023augmenting} introduced a method called Large-scale Language Models Augmented with medical textbooks (LLM-AMT). This method integrates medical textbooks as an external database for a question and answering system. Empirical evaluations suggested an enhancement in the accuracy of responses when utilizing LLM-AMT.

Although the approaches mentioned above have made notable strides in augmenting LLMs with external knowledge for the healthcare domain, they often rely on a single knowledge source or focus on specific medical tasks. However, in real-world healthcare scenarios, relevant information is typically distributed across multiple fragmented sources, such as patient medical records, clinical guidelines, research literature, and online knowledge bases. This data segmentation and diversity of knowledge sources pose a significant challenge in providing comprehensive and personalized medical responses tailored to a patient's unique context and history. As depicted in Fig. \ref{fig:our_rag}, our framework addresses this challenge by integrating information from multiple sources and contextualizing patients' details with relevant medical knowledge. This multi-source context augmentation approach enables \AlgoName to overcome the limitations of relying on a single knowledge source and provides personalized medical information that accounts for the fragmented nature of healthcare data.

\begin{figure*}[h]
    \centering
    \includegraphics[scale=.42]{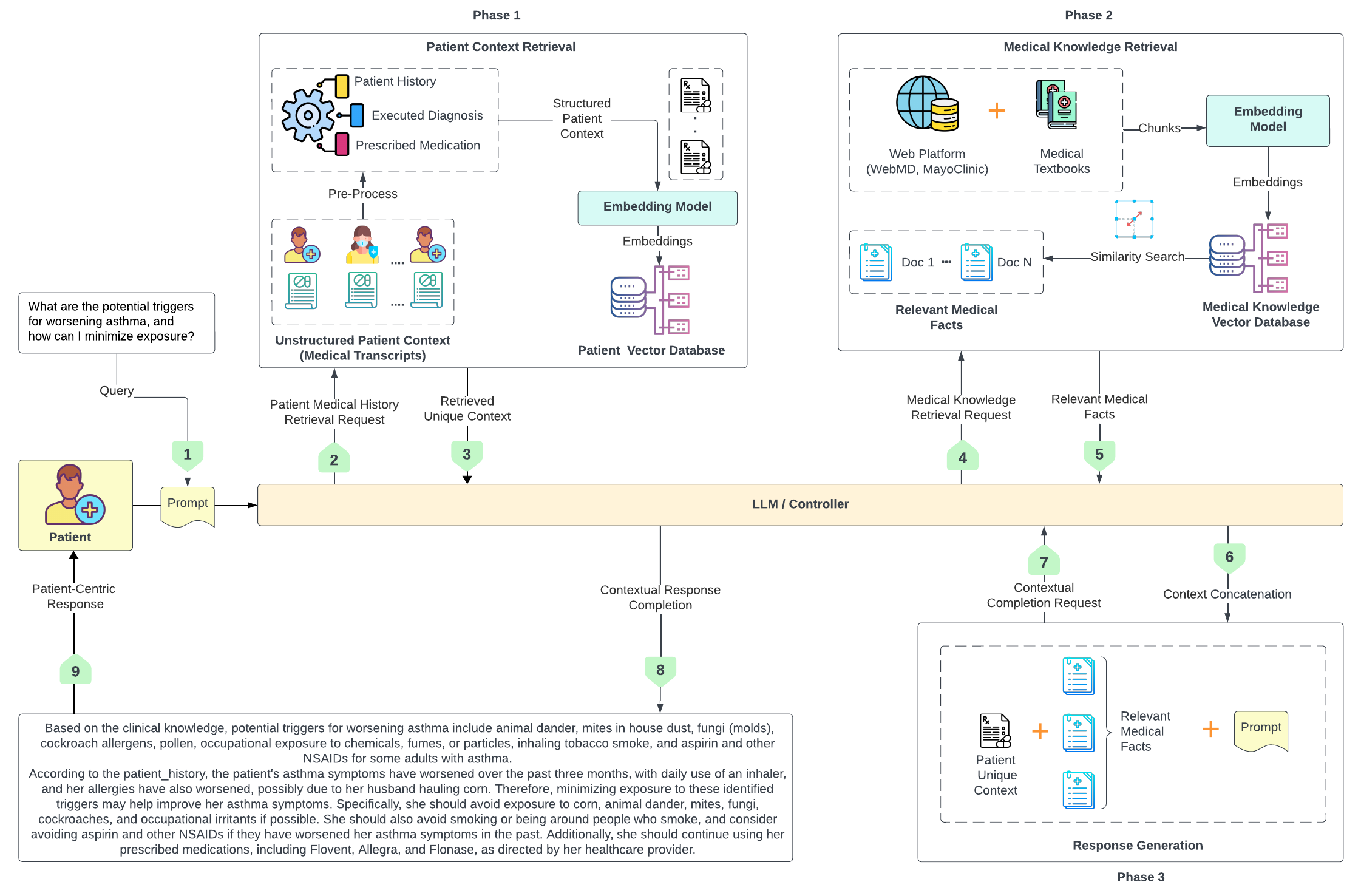}
    \caption{Detailed architecture of \AlgoName Framework}
        \label{fig: architecture}
\end{figure*}
\section{\AlgoName Architecture and Methodology} 
\label{architecture}

The architecture of our framework, referred to as \AlgoName, is delineated across three integral phases, each playing a distinct role in the functioning of the framework, as illustrated in Figure \ref{fig: architecture}. The three phases encompass \textit{patient context retrieval, medical knowledge retrieval} and \textit{response generation}. The initial phase revolves around patient context retrieval, specifically focusing on harnessing the patient's medical transcript to construct a comprehensive and patient-specific knowledge repository. 
In the subsequent phase, the framework extends its scope to encompass medical knowledge retrieval, leveraging trusted medical sources such as medical textbooks and web platforms (e.g., Mayo Clinic and WebMD). This phase ensures that our framework is well-versed in a variety of subject matter related to medical conditions, treatments, drugs, and general health and wellness. The final phase, \textit{response generation}, is the synthesis point where the accumulated contextual knowledge from both patient data or history 
and medical knowledge converges to craft a contextually relevant patient-centric response. This three-phase approach guarantees a robust and adaptive framework capable of delivering contextual and relevant information to meet the diverse needs of patients and caregivers.

In the following subsections, we explain each of these phases in greater details.

\subsection{Patient Context Retrieval}
In the healthcare domain, healthcare providers and patients engage in dialogues to address the health concerns of the patient. Subsequently, these conversations are transcribed to ensure the accuracy and comprehensiveness of patient records. Fig \ref{patient_context_transformation}(A) provides a glimpse of doctor-patient interaction. Within this context, a medical transcript can be defined as a written or typed document that meticulously records a patient's medical history, diagnosis, treatment, and pertinent details. These transcripts are crafted based on the notes taken by healthcare professionals, including doctors, nurses, or medical 
transcriber, and may be derived from recordings of the interactions that occur during a patient's engagement with the healthcare provider. A representative instance of a transcribed medical document is presented in Figure \ref{patient_context_transformation}(B). During this phase, we initiate the conversion of the unstructured medical transcript into a structured format. This transformation is accomplished through the application of a zero-shot prompting strategy utilizing OpenAI’s GPT-3.5-Turbo, managed through API calls. The process involves categorizing and annotating the unstructured medical transcript into three distinct categories: \textit{patient history and symptoms, executed diagnostics}, and \textit{prescribed medications} along with \textit{further instructions}. Figure \ref{patient_context_transformation}(C) provides an illustrative example of a pre-processed and annotated unique patient context resulting from this process. Further insights into the data pre-processing techniques and strategies employed for this undertaking are discussed in Section \ref{data_preprocessing}.

The second step in this phase is transformation of structured and annotated transcript into a vector database. This process relies on two crucial components: an embedding model and a vector database. Embedding refers to the technique used for representing natural language words and documents in a way that captures their meanings. This representation is typically a real valued vector in low-dimensional space. In our \AlgoName framework, the embedding model is used to vectorize and pre-process the patient's unique contexts to store them in a vector database. This allows performing a \textit{semantic search} for relevant knowledge retrieval. Similarly, A Vector Database (VD) is a specific type of database that stores data in the form of high-dimensional vectors. These vectors are mathematical representations derived from raw data, such as unstructured texts, while representing word-level semantic meaning. The advantage of a vector database is that it facilitates rapid similarity search and retrieval of data based on their vector distances, contrary to traditional databases that search based on exact matches or predefined criteria. In our \AlgoName framework, Chroma DB\footnote{Chroma, Website: trychroma.com}, an in-memory VD, allows storing and retrieving the most relevant pre-processed patient unique context based on the semantic or contextual meaning of the prompt.

\begin{tcbraster}[raster columns=3,raster equal height,nobeforeafter,raster column skip=2mm]

       \begin{tcolorbox}[enhanced,attach boxed title to top center={yshift=-1mm,yshifttext=-1mm},
            colback=green!10!white,colframe=gray!5!black,colbacktitle=gray!90!black, left=0.1mm, right=0.5mm, boxrule=0.50pt]
            \tiny
            {\fontfamily{qcr}\selectfont
        \textbf{Doctor}: Good afternoon. I've reviewed your medical history and the recent events that led to your referral. Let's discuss your symptoms and the results of the diagnostic tests.\\
\textbf{Patient}: Hello. Thank you for seeing me. It's been a bit concerning with the perioral swelling, especially after taking Keflex. \\
\textbf{Doctor}: I understand. Can you tell me more about the recent episode of perioral swelling? Did it occur immediately after taking Keflex, or was there a delay? ... \\
\textbf{Patient}: Yes, it happened shortly after taking Keflex. I hadn't experienced anything like that before.\\
\textbf{Doctor}: Given your history of grass allergies and environmental sensitivities, there's a possibility this could be an allergic reaction. We performed RAST allergy testing to identify specific food and environmental allergies. The results will help us better understand the triggers.}
        \textbf{(A)}
        \end{tcolorbox}
        \begin{tcolorbox}[enhanced,attach boxed title to top center={yshift=-1mm,yshifttext=-1mm},
            colback=green!10!white,colframe=gray!90!black,colbacktitle=gray!80!black, left=0.1mm, right=0.5mm, boxrule=0.50pt]
            \tiny
            {\fontfamily{qcr}\selectfont
A 34-year-old male presents today self-referred at the recommendation of Emergency Room physicians and his nephrologist to pursue further allergy evaluation and treatment.... admitted and treated and felt that his allergy reaction was to Keflex, which was being used to treat a skin cellulitis dialysis shunt infection...,  Strong for heart disease, carcinoma, and a history of food allergies, and there is also a history of hypertension.,CURRENT MEDICATIONS: , Atenolol, sodium bicarbonate, Lovaza, and Dialyvite.,ALLERGIES: , Heparin causing thrombocytopenia... blood pressure 128/78, pulse 70, temperature is 97.8, weight is 207 pounds, and height is 5 feet 7 inches...An EpiPen was also prescribed in the event of acute angioedema or allergic reaction...he is aware he needs to proceed directly to the emergency room for further evaluation and treatment recommendations after administration of an EpiPen.}
\textbf{(B)}
\end{tcolorbox}
\begin{tcolorbox}[enhanced,attach boxed title to top center={yshift=-1mm,yshifttext=-1mm},
            colback=green!10!white,colframe=gray!90!black,colbacktitle=gray!80!black, left=0.1mm, right=0.5mm, boxrule=0.50pt]
            \tiny
            {\fontfamily{qcr}\selectfont
\textbf{Patient history and symptom}:
- The patient is a 34-year-old male who was referred for further allergy evaluation and treatment after experiencing an acute event of perioral swelling of uncertain etiology, suspected to be a reaction to Keflex.
... food allergies, and hypertension.

\textbf{Executed diagnostics}:
- RAST allergy testing was performed to identify specific food and environmental allergies.
- The patient was advised to discontinue the use of cephalosporin antibiotics due to suspicion of an allergic reaction to Keflex.
- An EpiPen was prescribed for ...allergic reaction.

\textbf{Prescribed medications and further instructions}:
- Current medications include Atenolol, ...to prevent further reactions.
- In case of an allergic reaction, the patient should use the prescribed EpiPen and seek immediate medical attention for further evaluation and treatment.} \textbf{(C)}
\end{tcolorbox}

  
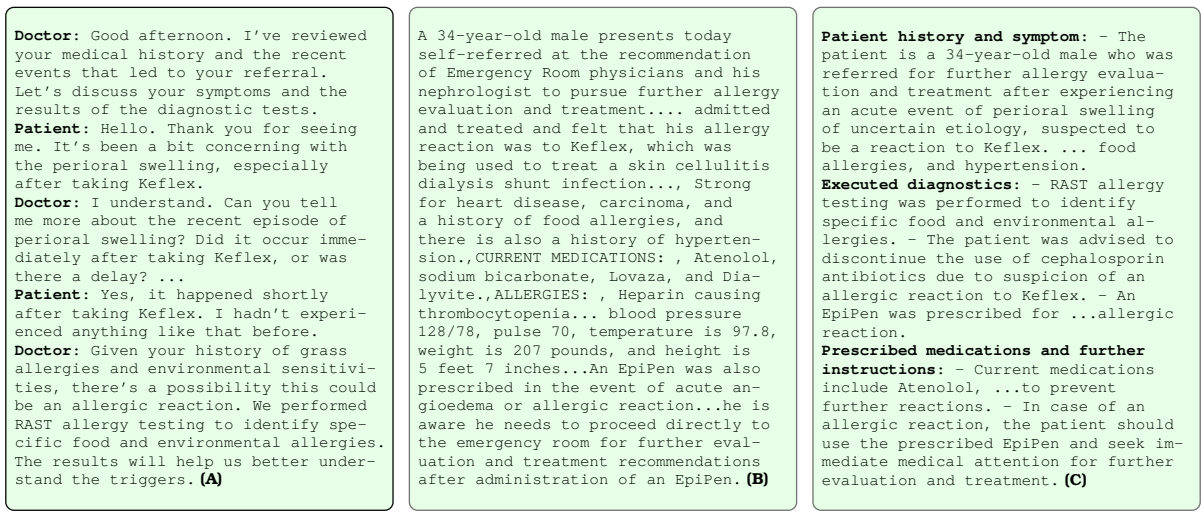
\captionof{figure}{Illustration the transformation of the doctor-patient interaction into the patient's specific context: \textbf{A} represents the interaction between the doctor and patient, \textbf{B} denotes an unstructured medical transcript, and \textbf{C} indicates the annotated structured representation of the patient's unique context.}
  \label{patient_context_transformation}

\end{tcbraster}

\subsection{Medical Knowledge Retrieval}
\label{medical_knowledge_retrieval}

A crucial phase in \AlgoName's framework is the comprehensive augmentation of the patient's context with relevant medical knowledge retrieved from authoritative sources like textbooks and trusted web platforms (e.g., Mayo Clinic, WebMD). Medical textbooks serve as authoritative repositories of factual information about conditions, symptoms, diagnoses, and treatments, while web platforms provide a diverse array of healthcare resources accessible to both professionals and the general public. This context augmentation phase takes the relevant chunks extracted from the patient's medical transcript in the initial phase and combines them with the user's query to establish a broader context. \AlgoName then leverages this comprehensive context to retrieve the most pertinent medical knowledge from the available sources.

\begin{center}
\begin{minipage}{14cm}
        \begin{tcolorbox}[enhanced,attach boxed title to top center={yshift=-1mm,yshifttext=-1mm},
            colback=green!10!white,colframe=gray!90!black,colbacktitle=gray!80!black, left=0.1mm, right=0.5mm, boxrule=0.50pt]
            \tiny
            {\fontfamily{qcr}\selectfont

            \textbf{Question}: How do \textcolor{red}{I} use the \textcolor{red}{prescribed EpiPen} in case of an emergency? \\ \\
            \textbf{Patient Unique Context}: \textbf{Patient history and symptom}:
                - The patient is a 34-year-old male who was referred for further allergy evaluation and treatment after...\textbf{Executed diagnostics}:- RAST allergy testing was performed to identify...An \textcolor{red}{EpiPen was prescribed}...
                \textbf{Prescribed medications and further instructions}:
- Current medications include Atenolol...the patient should use the \textcolor{red}{prescribed EpiPen.} \\ \\
\textbf{Medical Knowledge Retrieval}: \textbf{Document 1} The EpiPen or Anapen with 0.3 mg (300 $\mu$g) of adrenaline...\textcolor{red}{Place the orange tip against the middle}...\textcolor{red}{without bending or twisting it}..
\textbf{Document 2}:Indications for prescription of adrenaline for self-injection...\textbf{Document 3}: \textcolor{red}{Instruction regarding self-injection} of adrenaline...
            }
            
        \end{tcolorbox}
        \noindent\begin{minipage}{\textwidth}

\end{minipage}
\vspace{1mm}

\end{minipage}
\vspace{-8mm}
\captionof{figure}{An illustrative example of retrieving medical knowledge that is tailored to a patient's unique context when responding to a medical query.}\label{box_medical_knowledge_retrieval_with_context_}

\end{center}

To illustrate, Fig. \ref{box_medical_knowledge_retrieval_with_context_} provides an example of medical knowledge retrieval based on a patient's unique context and a specific medical query. In this particular scenario, the patient's unique context involves a male subject experiencing an allergic reaction, for which he is prescribed an EpiPen$^{\circledR}$ for emergencies. However, the instructions on how to use it are not specified in this patient's context. The medical knowledge retriever then searches for information on this topic and returns the most relevant documents based on the patient's context and medical query. The capability of MedInsight to search for pertinent information based on the provided context and query makes it a comprehensive and supportive technology for both patients and caregivers. The results are personalized, as evident in the figure, where a patient is seeking medical advice tailored to his specific context.

\begin{figure*}[h]
    \centering
    \includegraphics[scale=.85]{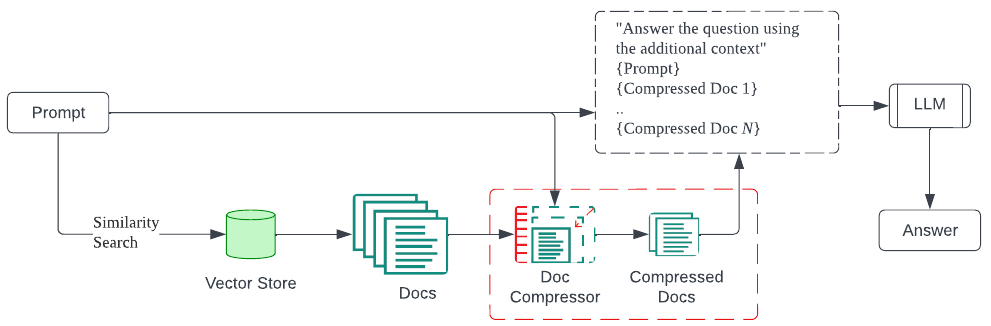}
    \caption{A graphical depiction showcasing contextually compressed RAG pipeline.}
        \label{fig: compressor_workflow}
\end{figure*}

However, traditional RAG pipelines can encounter challenges with context overflow when retrieving evidence from various sources. Retrievers often don't know the specific queries the document storage system will face, leading to situations where relevant information may be buried within irrelevant text. Passing full documents can be computationally expensive and may degrade performance. To address this, MedInsight employs contextual compression retrievers available through LangChain, as depicted in Fig \ref{fig: compressor_workflow}. Instead of returning entire documents, these retrievers compress the content using the query context, extracting only the most relevant portions. This contextual compression ensures the extraction of concise, relevant medical knowledge, facilitating the subsequent phase of patient-centric response generation in \AlgoName.


\subsection{Response Generation}

The conclusive phase in the \AlgoName framework is response generation, tasked with generating comprehensive, patient-centric responses by leveraging the augmented context obtained from multiple sources in the preceding phases. As discussed earlier, \AlgoName augments the patient's context extracted from their medical transcript by retrieving relevant medical knowledge from authoritative textbooks and trusted web platforms using contextual compression techniques. This context augmentation process ensures that the response generation phase has access to a comprehensive, personalized context that combines the individual's unique medical details with pertinent background information tailored to their specific condition and query.

Utilizing a LLM, this final phase tailors responses to individual patients or caregivers based on user prompts and the augmented context comprising retrieved medical documents and the patient's context. Our approach integrates both proprietary and open-source LLMs for response generation. Initially, we employ OpenAI's GPT-3.5-Turbo model to generate responses for 100 sets of questions (refer to Section \ref{dataset_description} for details on selected medical specialties and corresponding questions). Next, we employ the same pipeline to generate responses using Mistral-7B-Instruct, facilitating a comparative assessment of closed-source versus open-source models in patient-centric response generation tasks. The findings of this evaluation are presented in Section \ref{quantative_evaluation}. Fig. \ref{box_final_Completions} illustrates an instance of a patient-centric response generated by \AlgoName, addressing a medical query: \textit{``How do I use the prescribed EpiPen® in case of an emergency?''} The response incorporates the patient's medical context extracted from their transcript, relevant evidence retrieved from external knowledge sources based on that context, and the query itself. The contextual aspects integrated from multiple sources are highlighted in red for clarity.

\begin{center}
\begin{minipage}{14cm}
        
        \begin{tcolorbox}[enhanced,attach boxed title to top center={yshift=-1mm,yshifttext=-1mm},
            colback=green!10!white,colframe=gray!90!black,colbacktitle=gray!80!black, left=0.1mm, right=0.5mm, boxrule=0.50pt]
            \tiny
            {\fontfamily{qcr}\selectfont

            \textbf{Question}: How do \textcolor{red}{I} use the \textcolor{red}{prescribed EpiPen} in case of an emergency? \\ \\
            \textbf{Patient Unique Context}: \textbf{Patient history and symptom}:
                - The patient is a 34-year-old male who was referred for further allergy evaluation and treatment after...\textbf{Executed diagnostics}:- RAST allergy testing was performed to identify...An \textcolor{red}{EpiPen was prescribed}...
                \textbf{Prescribed medications and further instructions}:
- Current medications include Atenolol...the patient should use the \textcolor{red}{prescribed EpiPen.} \\ \\
\textbf{Medical Knowledge Retrieval}: \textbf{Document 1} The EpiPen or Anapen with 0.3 mg (300 $\mu$g) of adrenaline...\textcolor{red}{Place the orange tip against the middle}...\textcolor{red}{without bending or twisting it}..
\textbf{Document 2}:Indications for prescription of adrenaline for self-injection...\textbf{Document 3}: \textcolor{red}{Instruction regarding self-injection} of adrenaline...\\

                \textbf{Patient-Centric Response}: To use the \textcolor{red}{prescribed EpiPen} in case of an \textcolor{red}{emergency}... Make sure to remove the blue safety release cap, firmly press the orange tip against the outer thigh, hold it in place for a few seconds, and then seek immediate medical attention. 
            }

        \end{tcolorbox}
       \vspace{-14pt}
        \noindent\begin{minipage}{\textwidth}

\end{minipage}
\vspace{1mm}
\end{minipage}
\end{center}
\vspace{-5mm}
\captionof{figure}{Illustrating a patient-centric response generated by \AlgoName, this example highlights how the patient's context, extracted medical knowledge, and their medical query are contextualized, as indicated by the red highlights.}\label{box_final_Completions}


\section{Experimental Results and Discussion}
\label{experiment}
Our framework, \AlgoName, can function as a personalized support system for patients, aiding them in comprehending their medical conditions while also offering broader medical information on general topics. Additionally, caregivers can benefit from this system by posing personalized questions related to a patient, thereby optimizing caregiving. Here, we present a proof-of-concept experiment demonstrating the feasibility of our 
framework. 

\subsection{Dataset Preparation and Description}
\label{dataset_description}

Securing access to publicly available medical data poses a challenge due to the stringent privacy regulations imposed by HIPAA. Because of this, in this study, we leverage the medical transcripts sourced from MTSamples \cite{mtsamples} as the foundational dataset (or patient context) for our investigation. 

MTSamples dataset is fully synthetic, meaning it contains no actual patient information and has been artificially generated. This repository comprises transcriptions of 5000 medical reports spanning a diverse array of over 40 medical specialties. From this expansive range, we have meticulously chosen 10 specific specialties for focused analysis, including \textit{Allergy/Immunology, Pulmonary/Cardiovascular, Dermatology, Gastroenterology, General Medicine, Orthopedic, Neurology, Podiatry}, and \textit{Pediatrics – Neonatal} as depicted in Table \ref{tab:dataset_description}. Within each  category, we select 5 transcripts at random that represent patient's unique context. We then generate questions for each context, employing a zero-shot prompting strategy with \textit{GPT-3.5-Turbo}. This approach allows us to create a pair of synthetic questions for every distinct patient context. In total, we compile a collection of approximately 100 questions, each uniquely tailored to individual patients. Selected transcripts encompass a wide age spectrum, ranging from infants aged 2 months to adults aged 93 and they represent both gender. Inclusion of large age group and gender makes our dataset diverse this in turn, make \AlgoName comprehensive. 

For medical knowledge retrieval we carefully curated medical textbooks for each medical speciality to make our retriever comprehensive (see Section \ref{medical_knowledge_retrieval} for details). In total, we collect 12 medical textbooks in PDF format, with 10 directly related to specific medical specialties and 2 serving as medical encyclopedias. Notably, some of these books exceeded 8000 pages. The total token counts for the books are presented in Table \ref{tab:dataset_description}. For a detailed insight into pre-processing via the splitting and chunking strategies employed for these extensive texts please refer to Section \ref{implementation_detail}.

\subsubsection{Dataset Pre-processing}
\label{data_preprocessing}
The initial stage involves the preprocessing of synthetic raw data obtained from the MTSamples dataset. This raw data comprises summaries of patient contexts devoid of annotations. Through contextual analysis, we categorize the data into three distinct groups: patient history, executed diagnostics, and prescribed medications and further instructions. This categorization serves as a pivotal factor in guiding \AlgoName’s retrievers to precisely identify pertinent information within the text documents. Consequently, this approach facilitates the generation of contextually relevant, patient-centric responses. For the preprocessing of raw data, we employed a zero-shot prompting strategy, instructing the \textit{GPT-3.5-Turbo} model to annotate the provided patient context into the aforementioned three categories. Table \ref{table:annotation_data} illustrates the example of patient context and it's annotated output after applying zero-shot prompting techniques. 

\begin{table}
 \def\arraystretch{1.3}
 \footnotesize

     \caption{Medical speciality dataset curated from MTSAmples.}\label{tab:dataset_description}
   \begin{tabular}{ *{4}{c} }

     \hline
     \textbf{Medical Speciality} & \textbf{\# of Transcript(s)} & \textbf{\# Selected}  & \textbf{\# of Token(s) }  \\\hline

          Allergy / Immunology & 7 & 5 & 335,807 \\ 
      Pulmonary / cardiovascular & 372&  5 & 643,733\\
       Dermatology & 29 & 5 & 652,882 \\
      Gastroenterology & 230 & 5 & 108,893 \\
       General Medicine & 259 & 5 & 541,243\\
     Orthopedic & 355& 5 & 481,512 \\
     Neurology & 223&  5 & 1,921,963\\
     Podiatry & 47& 5 & 82,270\\
     Urology & 158& 5 & 791,163  \\
      Pediatrics - Neonatal & 70 & 5  & 1,356,563  \\
      \bottomrule
      Total & 1750 & 50 & 6,263,147 \\
  
    \end{tabular}
    \end{table}
   
    

\subsection{Implementation Details}
\label{implementation_detail}
In the course of our experiments, we employed OpenAI's proprietary GPT-3.5-turbo and the open-source Mistral-7B-Instruct as the foundational generator for our framework. The model's temperature was intentionally set to 0 to eliminate randomness in the response. GPT-3.5-turbo, accessed through its API, handled the patient context annotation task using zero-shot prompting strategies. For both patient context retrieval and medical knowledge retrieval, we constructed a vector database using Chroma 
however, the chunking and splitting strategies differed. The patient context, being relatively smaller in size, was split into 500 chunks with an overlap of 200. In contrast, the medical context, curated from web platforms and medical textbooks, was significantly larger, with some medical books comprising more than 8 thousand pages. Hence, we opted for a chunk size of 2500 with a 500 overlap for the medical knowledge retriever. The base embedding model in our RAG pipeline is \textit{text-embedding-ada-002}. During the evidence retrieval stage, we utilized the \textit{contextual compression retriever} available through LangChain. Instead of immediately returning retrieved documents as-is, this retriever compresses them using the context of the given query, ensuring that only relevant information is returned. Such retrievers enhance the efficiency and effectiveness of the document retrieval process, resulting in better user experiences and optimized resource utilization. The generation of contextual, patient-centric responses involved both GPT-3.5-Turbo and Mistral-7B-Instruct 8 Bit quantized models. The experiment was conducted in the Google Colab Jupiter environment with standard CPU runtime for GPT-3.5-Turbo. Mistral-7B-Instruct was downloaded from Huggingface\cite{mistral} and run locally. We utilized an {Intel i9-12900 CPU}, GPU GeForce RTX™ 3090 Ti with 24 GB and 128 GB of RAM to run Mistral. High level overview of our algorithm is provided in Algorithm \ref{algorithm_MedInsight}.

\begin{algorithm}
\footnotesize
\SetAlgoNlRelativeSize{0}
\KwInput{Prompt $p$ }
\KwRequire{a patient context $D_p$, a patient context retriever $R_p$, medical knowledge $D_m$ , a medical context retriever $R_m$, retrieval augmented generator $G$}
\textbf{Variables:} $D_pR_p$ is retrieved patient context,  $D_mR_m$ is retrieved medical knowledge \\
 \KwOutputput{Patient-centric response ($C$)}
 1: Request $D_p$ with prompt $p$ \;
 2: Retrieve $D_p$ with patient retriever $R_p$ ->Return relevant patient context $D_pR_p$\;
 3: Request $D_m$ with prompt $p$\;
 4: Retrieve $D_m$ with medical retriever $R_m$ -> Return relevant medical knowledge $D_mR_m$\;
 5: Concatenate $D_pR_p$ $\oplus$ $D_mR_m$ $\oplus$ $p$\;
 6: Use $G$ to generate patient-centric response using concatenated context -> $G $($D_pR_p$ $\oplus$ $D_mR_m$ $\oplus$ $p$)$ $ \;
 7: return $C$
\caption{\AlgoName Framework}
\label{algorithm_MedInsight}
\end{algorithm}


\subsection{Evaluation}
 
   



\begin{table*}
 \caption{Example highlighting the transformation of unstructured patient context into a structured format using a zero-shot prompting strategy. Due to space limitations, the table presents only a snapshot of the actual data.}
    \label{table:annotation_data}
    \vspace{-1mm}
\footnotesize
\centering

{\renewcommand{\arraystretch}{1.30}%
    
    \begin{tabularx}{\textwidth}{XXX}
     
    \hline
    \textbf{Prompt Template} & 
    \textbf{Unstructured Patient Context} & 
    \textbf{Annotated Unique Patient Context}\\
    \hline
   
    prompt\_template = """Given the following \{medical transcript\} of a patient, create a detailed summary by categories. The summary should be divided into the following categories:
    
    \begin{minipage}[t]{\linewidth}
    \begin{itemize}[leftmargin=*]
        \item  Patient history and symptom
        \item Executed diagnostics
        \item  Prescribed medications \& Instruction:
    \end{itemize}
        Medical Transcript:
    \{transcript1.txt\}
    \end{minipage} """  &
    
    An 83-year-old diabetic female presents today stating that she would like diabetic foot care.,O - ,On examination, the lateral aspect of her left great toenail is deeply ingrown.  Her toenails are thick and opaque.  Vibratory sensation appears to be intact.  Dorsal pedal pulses are 1/4.  There is no hair growth seen on her toes, feet or lower legs.  Her feet are warm to the touch...discolored &

    \begin{minipage}[t]{\linewidth}
    \begin{itemize}[leftmargin=*]
        \item Patient history and symptom: The patient is an 83-year-old diabetic female presenting with a complaint of diabetic foot care...discolored.
        \item Executed diagnostics: ...the diagnosis made is onychocryptosis,... 
        \item Prescribed medications \& Instruction: The transcript does not ... prescribed medications...
    \end{itemize} 
    \vspace{1mm}
    \end{minipage} \\
  
    \hline 


   prompt\_template = """Given the following \{medical transcript\} of a patient, create a detailed summary by categories. The summary should be divided into the following categories:
    
    \begin{minipage}[t]{\linewidth}
    \begin{itemize}[leftmargin=*]
        \item  Patient history and symptom
        \item Executed diagnostics
        \item  Prescribed medications \& Instruction:
    \end{itemize}
        Medical Transcript:
    \{transcript2.txt\}
    \end{minipage} """  &
    
    The patient is admitted for shortness of breath, continues to do fairly well.  The patient has chronic atrial fibrillation, on anticoagulation, INR of 1.72...cardiologist regarding aortic stenosis.  She may need a surgical intervention in this regard, which I explained to her.  The patient will be discharged home on medical management and she has an appointment to see her cardiologist in the next few days.,In the interim, if she changes her mind or if she has any concerns, I have requested to call me back.&

    \begin{minipage}[t]{\linewidth}
    \begin{itemize}[leftmargin=*]
        \item Patient history and symptom: The patient was admitted for shortness of breath...atrial fibrillation and is on anticoagulation with an INR of 1.72...severe aortic stenosis...
        \item Executed diagnostics: Physical examination showed vital signs...systolic murmur in the aortic area,...impression was made of shortness...
        \item Prescribed medications \& Instruction: Continue current medications...cardiologist regarding aortic stenosis, as surgical intervention may be necessary.

    \end{itemize} 
    \vspace{1mm}
    \end{minipage} \\
  
    \hline

    \end{tabularx}
   
}
\end{table*}

Our evaluation study encompasses a dual-pronged experimental approach. For the quantitative assessment of our framework's performance in generating contextually relevant responses, we employ a comprehensive set of metrics, including Ragas scores \cite{es2023ragas} and TruLens scores \cite{TruLens}. 
Following the quantitative assessment of our model's overall effectiveness, we proceeded with a Subject-Matter Expert (SME) evaluation study to validate the answer generation capabilities of \AlgoName. Due to the resource-intensive nature of this evaluation, we engaged a panel of four medical residents. 
Their task involved scoring answers across 100 questions spanning across all medical speciality as explained in Section \ref{dataset_description}, considering two critical aspects: factual correctness and relevance to the patient’s unique context.

\subsubsection{Quantitative Evaluation}
\label{quantative_evaluation}

To quantitatively evaluate the performance of our \AlgoName in generating contextually relevant patient-centric responses, we utilize RAGAS~\cite{es2023ragas} and TruLens \cite{TruLens} frameworks. These frameworks feature 
offers comprehensive metrics specifically designed for assessing RAG pipelines. We chose these frameworks 
over popular alternatives like BLEU \cite{papineni2002bleu} and ROUGE \cite{rouge2004package}, as they are not aligned with our specific context. BLEU is primarily used to evaluate machine translation tasks, while ROUGE is specialized for evaluating text summarization tasks. Both metrics focus on structural similarity between ground truth and generated sentences, which may not be suitable for our case where sentences can be structurally different but factually similar. Traditional metrics fail to capture this nuance. Given that \AlgoName predominantly functions as a RAG-based question-answering system with context mapping, these conventional metrics are not well-suited to gauge its effectiveness.
\begin{table}[h]
\def\arraystretch{1.3}
 \caption{Quantitative evaluation using both proprietary and open source LLM for answer generation task. In our context, gpt-3.5-turbo has slight edge on answer similarity. While it has over 7\% gain on answer correctness over Mistral-7B model.}
    \footnotesize

 \label{quantative_table}

{\renewcommand{\arraystretch}{1.3}%
\centering
\begin{tabular}{l|l|l|l}
\hline
\textbf{Evaluation   Framework}   & \textbf{Model}               & \textbf{Average   Similarity Score} & \textbf{Average   Correctness Score} \\ \hline
\multirow{2}{*}{Ragas}   & gpt-3.5-turbo       & \cellcolor{green!15}0.93                       & \cellcolor{green!15}0.84                        \\ 
                         & Mistral-7B-Instruct & 0.92                       & 0.77                        \\\hline
\multirow{2}{*}{TruLens} & gpt-3.5-turbo       & 0.90                        & -                       \\ 
                         & Mistral-7B-Instruct & 0.91                       & -      \\
                         \hline
        
\end{tabular}}

\end{table}

\begin{figure}
 \begin{subfigure}{0.49\textwidth}
     \includegraphics[width=\textwidth]{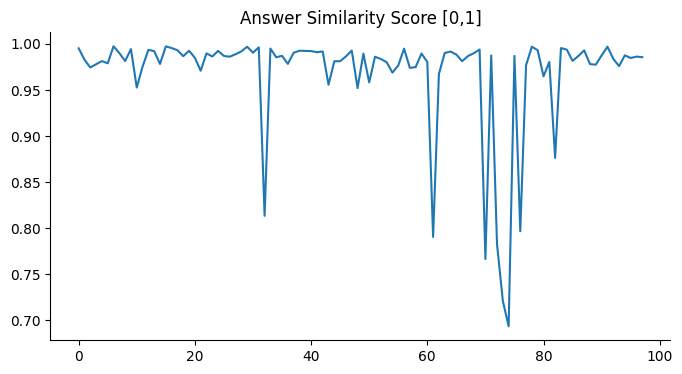}
     \caption{Evaluation using RAGAS for GPT-3.5-Turbo answer generated }
    \label{fig:Ragas_GPT_answer_similarity}
 \end{subfigure}
 \hfill
 \begin{subfigure}{0.49\textwidth}
     \includegraphics[width=\textwidth]{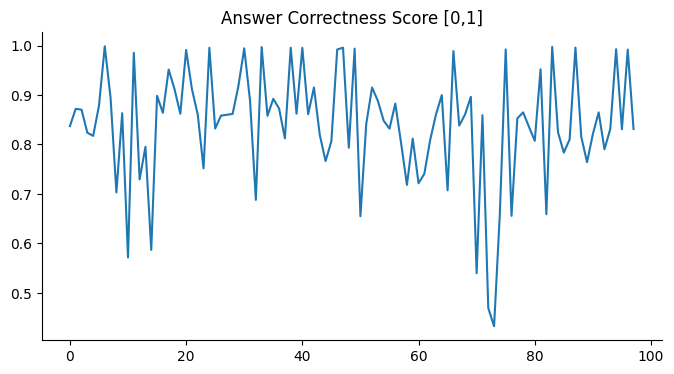}
     \caption{Evaluation using RAGAS for GPT-3.5-Turbo answer generated on answer correctness.}
    \label{fig:Ragas_GPT_answer_correctness}
 \end{subfigure}
 
 \medskip
 \begin{subfigure}{0.49\textwidth}
     \includegraphics[width=\textwidth]{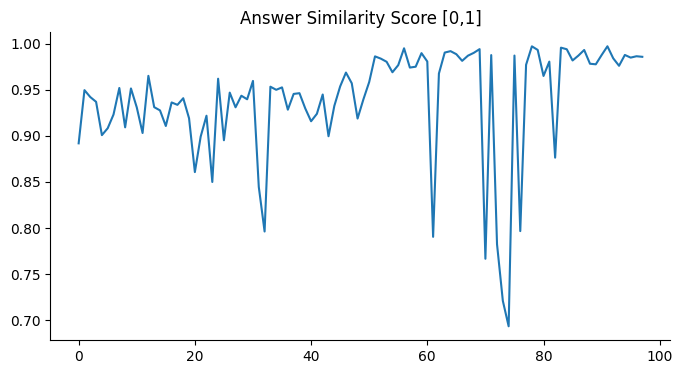}
     \caption{Evaluation using RAGAS for Mistral-7B-Instruct answer generated on answer similarity.}
    \label{fig:Mistral_answer_similarity}
 \end{subfigure}
 \hfill
 \begin{subfigure}{0.49\textwidth}
     \includegraphics[width=\textwidth]{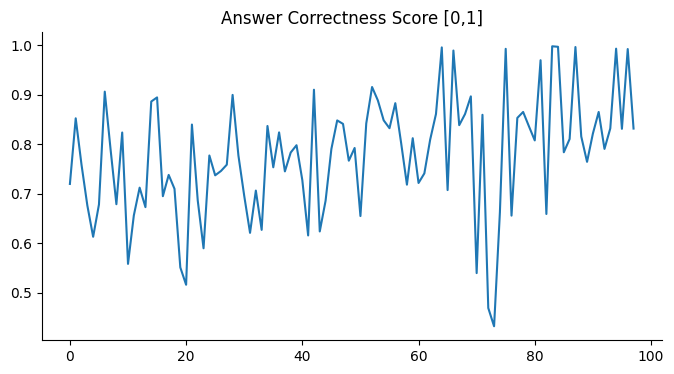}
    \caption{Evaluation using RAGAS for Mistral-7B-Instruct answer generated on answer correctness.}
    \label{fig:Mistral_answer_correctness}
 \end{subfigure}
 \caption{Quantitative assessment highlighting the average RAGAS scores for both answer similarity and correctness, comparing the performance of GPT-3.5-turbo and Mistral-7B-Instruct.}
 \label{Label}

\end{figure}

In the context of our study, we report the mean RAGAS score of \textit{0.93} for answer similarity and \textit{0.8409} for answer correctness in case of GPT-3.5-turbo generated answers. Figure \ref{fig:Ragas_GPT_answer_similarity} depicts answer similarity scores, and Figure \ref{fig:Ragas_GPT_answer_correctness} presents answer correctness scores for answers generated by GPT-3.5-Turbo. Whereas, in the case of Mistral-7B-Instruct, we report a mean score of \textit{0.92} for answer similarity and \textit{0.77} for answer correctness. Figure \ref{fig:Mistral_answer_similarity} illustrates the answer similarity score, while Figure \ref{fig:Mistral_answer_correctness} presents the answer correctness scores for responses generated by Mistral-7B-Instruct. Similarly, for TruLens, we report an average answer similarity score of \textit{0.91} for GPT-3.5-Turbo generated answers and \textit{0.90} for Mistral-7B generated answers. For both frameworks the scores range between \textit{0} and \textit{1}, where \textit{1} signifies optimal generation.

The results as highlighted in Table \ref{quantative_table} suggest that the GPT model exhibited a slight advantage over Mistral-7B-Instruct in terms of answer similarity. Additionally, it achieved a notable 7\% increase in accuracy when assessed in relation to answer correctness. Overall, results emphasize the proficiency of \AlgoName\ in delivering contextually relevant answers within the framework of retrieval-augmented question-answering tasks.

\subsubsection{Qualitative Evaluation}
After evaluating the effectiveness of our model through quantitative metrics, we proceed to a human evaluation study to validate \AlgoName's proficiency in the patient-centric response generation task. Given the substantial cost associated with this evaluation, we engage a panel of 4 medical professionals, comprising medical doctors and residents, to evaluate the generated responses using scoring mechanism. Evaluation is conducted on two key aspects: factual correctness and contextual relevancy. The first criterion measures the accuracy and relevance of the generated answer to the given question, while the second criterion assesses the contextual appropriateness of the retrieved information, taking into account both the question and the patient's unique context.

Medical experts were provided with 10 distinct medical specialties, each comprising 5 unique patient contexts and 2 question-answer pairs (see Section \ref{dataset_description} for details about questions). Their assignment was to manually evaluate the quality of answers generated across a total of 100 question-answer pairs. To evaluate factual correctness and contextual relevance, a 5-point Likert scale \cite{allen2007likert} was employed, ranging from 1 (indicating "Factually Incorrect and Contextually irrelevant") to 5 (indicating "Factually Accurate and Contextually relevant"). Medical professionals were instructed to score responses on this scale. The average score obtained was 4.66 out of 5. This step allowed us to establish the ground truth for quantitative evaluation. Additionally, we evaluated inter-rater agreement as depicted in Table \ref{tab:felis} using the Fleiss Kappa measure \cite{mchugh2012interrater}.The evaluation revealed a moderate overall agreement among medical experts on the generated responses, standing at 0.60 with a standard error of 0.029.

\begin{table}
\caption{Fleiss Multirater Kappa Analysis: Moderate agreement (0.60) among four raters evaluating 98 sets of answers across 10 medical specialties.} \label{tab:felis}
 \vspace{-1mm}

    {\renewcommand{\arraystretch}{1.3}%
\footnotesize

\begin{tabular}{ll|l} 
\hline
                                    & \multicolumn{2}{l}{\textbf{Overall Categories $^ {a,b}$}}  \\ \cline{1-3}
\multirow{2}{*}{Overall Agreement}  & Kappa  ($K$)             & Standard Error        \\ \cline{2-3}
                                    & \cellcolor{green!15} 0.600708945         & 0.029630951           \\ \cline{1-3}
\multicolumn{3}{l}{a. Data contains 98 question-answer pairs evaluated by 4 raters using Fleiss Kappa.} \\
\multicolumn{3}{l}{b. Rating category values are case-sensitive.}              
\end{tabular}

}
 
\end{table}

\section{Conclusion and Future Work}

Large Language Models possess remarkable capabilities in generating contextual responses. However, their application in the healthcare domain is limited due to deficiencies in domain-specific knowledge. To address this, we have developed a novel framework called \AlgoName, aiding patients in better understanding their medical history, diagnosis, and prescriptions through retrieval-augmented question-answering. The main objective of \AlgoName is to empower patients with insights for improving and optimizing both patient care and healthcare delivery.

\indent To achieve this, \AlgoName employs a context augmentation approach that combines medical knowledge from multiple sources like medical textbooks and web platforms with patients' unique medical context from their transcripts. The developed method comprises three stages: First, relevant details from medical transcripts and health records are extracted to understand the patient's context. Second, MedInsight retrieves trusted and relevant clinical information from external resources like WebMD and Mayo Clinic to augment the patient's context. Finally, the augmented context, encompassing the patient's details and retrieved medical knowledge, is utilized to generate patient-centric responses to the user prompt.

We evaluated the effectiveness of \AlgoName's context augmentation approach using the MTSamples dataset with ten medical conditions and fifty unique patient contexts. Results demonstrated \AlgoName's efficacy in generating contextually relevant responses. The RAGAS framework revealed promising scores for answer similarity (0.93 for GPT-3.5-turbo and 0.92 for Mistral-7B-Instruct) and answer correctness (0.84 for GPT-3.5-turbo and 0.77 for Mistral-7B-Instruct). 

In the future we aim to further optimize the retrievers and investigate on effectiveness of RAG approach over fine-tuning for augmented context from multiple source when generating contextually relevant patinet-centric responses. 

\label{conclusion}

\section*{Acknowledgments}

This work was supported by PATENT Lab (Predictive Analytics and TEchnology iNTegration Laboratory) at the Department of Computer Science and Engineering, Mississippi State University. The authors would like to thank SME's for their assistance in qualitative evaluation. The views and conclusions are those of the authors.


\bibliographystyle{unsrt}
\bibliography{sample-base}


\end{document}